\documentclass{article}
\usepackage{spconf,amsmath,graphicx,hyperref}

\usepackage{multirow}
\usepackage{xcolor}
\usepackage{booktabs}
\usepackage{amsmath}
\usepackage{cite}

\title{MUGSQA: Novel Multi-Uncertainty-Based Gaussian Splatting Quality Assessment Method, Dataset, and Benchmarks}
\name{Tianang Chen,
    Jian Jin*,
    Shilv Cai,
    Zhuangzi Li,
    Weisi Lin*\thanks{* Corresponding author.}}
\address{Nanyang Technological University}
\begin{document}
\ninept
\maketitle

\begin{abstract}
Gaussian Splatting (GS) has recently emerged as a promising technique for 3D object reconstruction, delivering high-quality rendering results with significantly improved reconstruction speed. As variants continue to appear, assessing the perceptual quality of 3D objects reconstructed with different GS-based methods remains an open challenge. To address this issue, we first propose a unified multi-distance subjective quality assessment method that closely mimics human viewing behavior for objects reconstructed with GS-based methods in actual applications, thereby better collecting perceptual experiences. Based on it, we also construct a novel GS quality assessment dataset named MUGSQA, which is constructed considering multiple uncertainties of the input data. These uncertainties include the quantity and resolution of input views, the view distance, and the accuracy of the initial point cloud. Moreover, we construct two benchmarks: one to evaluate the robustness of various GS-based reconstruction methods under multiple uncertainties, and the other to evaluate the performance of existing quality assessment metrics. Our dataset and code are available at \url{https://github.com/Solivition/MUGSQA}.
\end{abstract}


\begin{figure*}[t]
    \centering
    \small
    \includegraphics[width=0.9\linewidth]{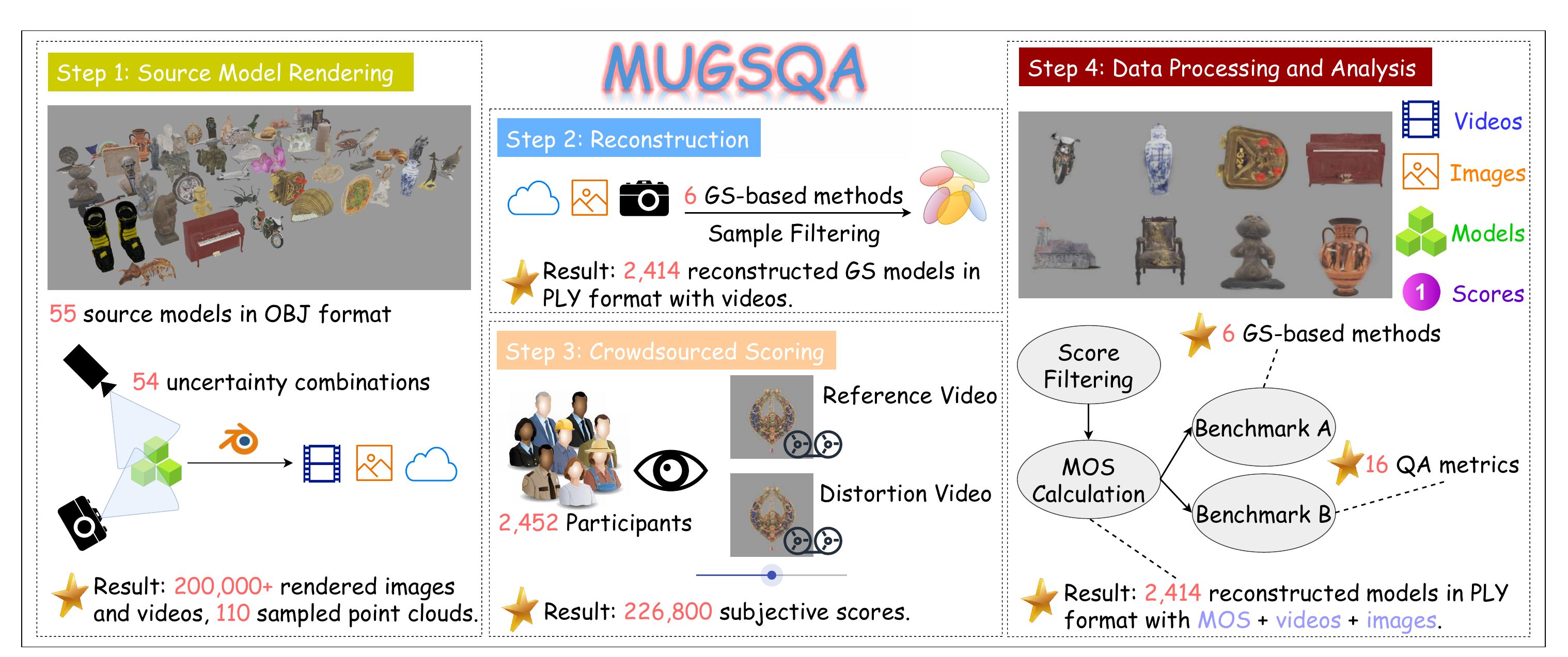}
    \caption{MUGSQA. In Step 1, we select 55 source models, render, and sample on them. During this process, we simulate a total of 54 combinations of uncertainties that might cause differences. In Step 2, we first employ 6 GS-based methods to reconstruct these models. Then, we render all samples and their source models into videos and filter them according to their quality. In Step 3, we utilize these videos and our SQA method to collect quality scores during subjective experiments. In Step 4, we filter the scores and complete the dataset. Finally, we construct two benchmarks aimed at evaluating existing metrics and comparing the robustness of different GS-based reconstruction methods.}
    \label{fig:overall_pipeline}
\end{figure*}

\section{Introduction}
\label{sec:intro}
3D reconstruction is a fundamental problem in computer vision, aiming to recover accurate geometry and appearance of real-world objects and scenes. Among emerging approaches, the first method based on Gaussian Splatting (GS) \cite{kerbl20233d} offers a compelling balance between high rendering quality and real-time performance. Its outstanding performance quickly makes it one of the most promising solutions for practical deployment in 3D object reconstruction and draws attention from both academia and industry.

Although numerous GS-based reconstruction methods \cite{kerbl20233d, fan2024lightgaussian, yu2024mip, lu2024scaffold, girish2024eagles, ren2024octree} have recently been proposed, two fundamental questions remain underexplored: i) How well can GS-based reconstruction methods sustain their performance under different input \textbf{uncertainties} \cite{klasson2024sources} (\emph{e.g.,} different numbers of input views, different initial point clouds, and so on)? ii) Are existing quality assessment metrics \cite{zhang2018unreasonable, zhang2020blind, wu2024qalign} adequate for evaluating such methods? These questions are pivotal not only for enabling fair comparisons among competing methods but also for driving the continuous improvement of reconstruction performance. To answer the above questions, the benchmarks for GS Quality Assessment (GSQA) are required.

Existing quality assessment benchmarks have primarily focused on images \cite{hosu2020koniq, ying2020patches}, point clouds \cite{yang2020predicting, liu2023point}, and meshes \cite{nehme2023textured, cui2024sjtu}. Only a few studies \cite{yang2024benchmark, zhang2025evaluating, martin2025gs, xing20253dgsieval15klargescaleimagequality},  have constructed GSQA datasets, but these works mainly target compression-induced degradations \cite{yang2024benchmark, xing20253dgsieval15klargescaleimagequality}, rather than the more common distortions arising from input uncertainties during GS reconstruction. Such uncertainties include failed occlusion recovery under sparse view density, detail loss due to low-resolution inputs, perspective distortion from changes in view-to-object distance, and structural deviations caused by inaccuracies in the initial point cloud. Consequently, current GSQA datasets are insufficient not only for comprehensive benchmarking of GS-based reconstruction methods, but also for validating the effectiveness of existing quality metrics in capturing distortions induced by these uncertainties. This limitation has further led to stagnation in the development of the GSQA metric design. To address this gap, we systematically introduce \textbf{M}ultiple \textbf{U}ncertainties during the data preparation process, adopt various \textbf{GS}-based reconstruction methods, and construct a new \textbf{Q}uality \textbf{A}ssessment dataset, termed \textbf{MUGSQA}. Unlike prior work that relies on real-world 2D captures, we select OBJ-format mesh models as reconstruction sources \cite{nehme2023textured, ma2025shapesplat}. By focusing on single-object scenes, our dataset eliminates interference from multiple coexisting objects, making it more suitable for controlled distortion analysis and metric design.

Besides, existing Subjective Quality Assessment (SQA) methods often present the 3D object to the subjects with a fixed view or a single-distance display \cite{zhang2025evaluating, martin2025gs}, making it difficult to reflect the behavior of subjects when dynamically observing Gaussian objects \cite{yang2024gaussianobject} in interactive or immersive scenarios. In order to better align with the above, we propose a unified multi-distance SQA method that guides observers to examine Gaussian objects from various distances and multiple views. Based on this, we conduct a large-scale subjective experiment to collect quality scores for the MUGSQA dataset. We gather 2,452 participants and ultimately obtain over 226,800 valid scores, ensuring that the scores we finally collect are sufficient and reliable. Finally, we construct two benchmarks based on the MUGSQA dataset to evaluate the robustness of GS-based reconstruction methods and the performance of existing objective quality assessment metrics on Gaussian objects. This fills the gap in the current evaluation system in this field and promotes the standardized development of GSQA.

The overview of MUGSQA is shown in Figure \ref{fig:overall_pipeline}. In summary, our main contributions include the following points:
\begin{itemize}
    \item We propose a unified multi-distance SQA method for Gaussian objects to capture the real subjects' quality experience.
    \item We construct the MUGSQA, which is a large-scale Gaussian object dataset taking into account different uncertainties and various GS-based reconstruction methods.
    \item We construct a benchmark on MUGSQA to evaluate the reconstruction robustness of representative GS-based methods under diverse uncertainties.
    \item We construct a benchmark on MUGSQA to evaluate the performance of existing quality assessment metrics for GSQA.
\end{itemize}

\section{Data Preparation}
\label{sec:preparation}

\begin{figure*}[t]
    \centering
    \includegraphics[width=0.85\linewidth]{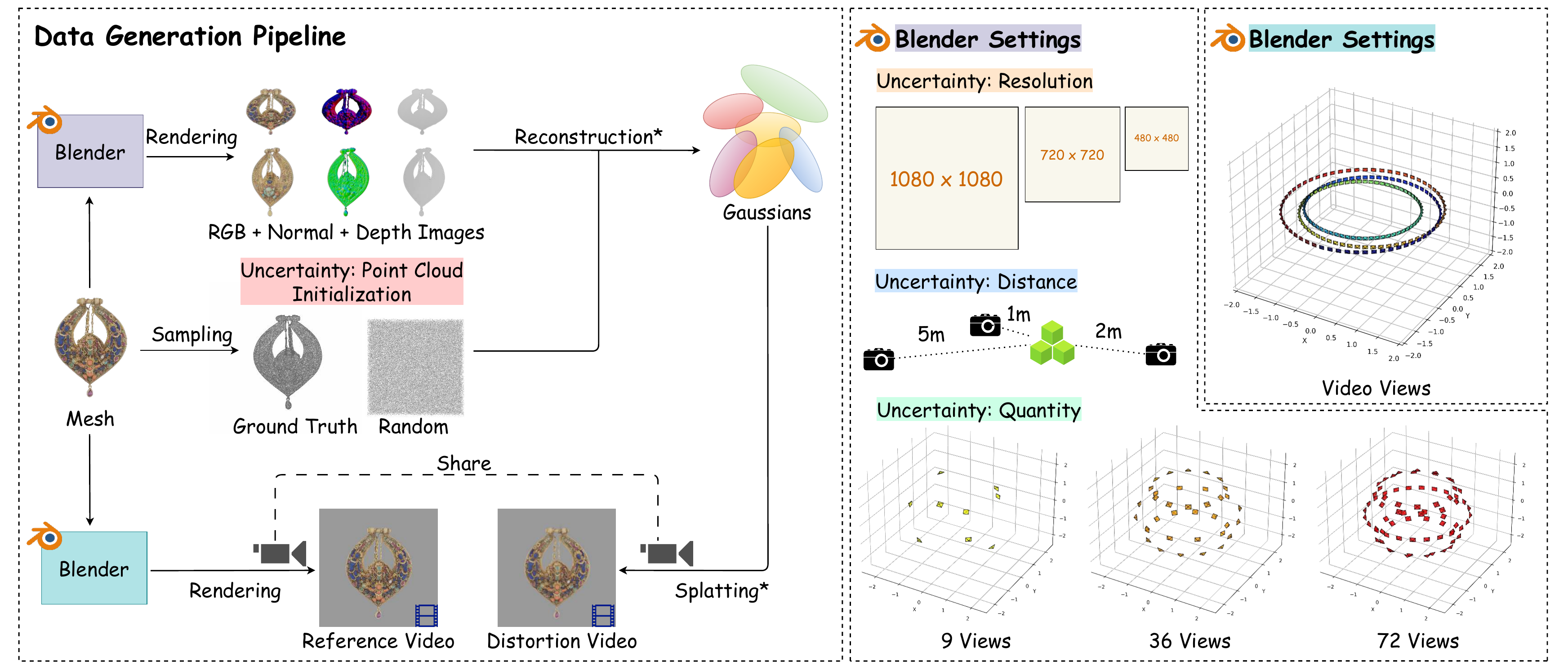}
    \caption{Data Generation Pipeline. From left to right, the first part represents the process of generating distorted samples and SQA videos; the second and third parts represent the reconstruction input uncertainty rendering settings and the rendering settings for SQA videos in Blender, respectively. The ``Share'' in the figure indicates the use of the same camera parameters. The ``Reconstruction*'' and ``Splatting*'' steps in the figure represent the use of the corresponding algorithm based on the selected GS-based reconstruction method.}
    \label{fig:pipeline}
\end{figure*}

\textbf{Source Models.} We select 55 mesh models as ground truth from Sketchfab\footnote{https://sketchfab.com/features/free-3d-models}, which have been demonstrated to have high geometric complexity and high texture quality \cite{nehme2023textured}.

\textbf{Main Set.} To obtain the input data required for reconstruction, we first render multi-view images in Blender\footnote{https://www.blender.org}, and export them together with camera poses and point clouds in the NeRF Synthetic Dataset \cite{mildenhall2021nerf} format. In accordance with \cite{ma2025shapesplat}, we do not use 3DGS \cite{kerbl20233d} but instead use LightGaussian \cite{fan2024lightgaussian} for reconstruction, ensuring that overall reconstruction quality does not suffer. To generate stimuli of different qualities, we simulate multiple uncertainties that may be encountered during actual data preparation. In our setup, the size of all objects is normalized, so our parameters must meaningfully reflect how the reconstruction degrades on this scale. Inspired by \cite{klasson2024sources}, we use the following settings in Blender: \textbf{(1) View resolution settings:} We choose \(1080\times1080\), \(720\times720\), and \(480\times480\) to model different observations. \textbf{(2) View quantity settings:} We use 72, 36, and 9 views. Here, 72 views ensure dense sampling with minimal occlusion, 36 aligns with standard multi-view datasets \cite{xu2019disn}, and 9 simulates realistic sparse-view conditions. The specific positions of the three quantities of views are shown in Figure \ref{fig:pipeline}. \textbf{(3) View-to-object distance settings:} 5\,m, 2\,m, and 1\,m correspond respectively to far-range overview, mid-range balanced capture, and close-up focus. \textbf{(4) Point cloud initialization settings:} Randomly sample \(10^5\) point clouds from either the model surface or the full scene, which allows us to simulate ideal initialization versus noisy inputs. These values are carefully chosen to match the unit-scale object space and cover a wide range of common distortion factors. They define a well-controlled and representative space for evaluating reconstruction algorithms under varied and realistic degradation conditions. Furthermore, to ensure that the quality distribution of the dataset falls within a common range of distortion, we perform data filtering to exclude samples that completely fail to reconstruct. Finally, our MUGSQA dataset contains 1,970 main set samples.

\textbf{Additional Set.} Next, we construct an additional set using more reconstruction methods. For this set, we only use 3 out of the 55 source models, but employ 5 GS-based methods for reconstruction: 3DGS \cite{kerbl20233d}, Mip-Splatting \cite{yu2024mip}, Scaffold-GS \cite{lu2024scaffold}, EAGLES \cite{girish2024eagles}, and Octree-GS \cite{ren2024octree}. This choice is made to keep the subjective experiment manageable while still covering diverse geometric and textural characteristics. All other settings remain consistent with the main set. Similarly, we filter this set and obtain 444 additional set samples. In total, we obtain \(1,970+444=2,414\) reconstructed models. Figure \ref{fig:pipeline} shows the overall data generation pipeline.

\section{Subjective Quality Assessment}
\label{sec:sqa}
\textbf{Method.} To fully assess the quality of Gaussian objects, we propose a unified multi-distance SQA method. As shown in Figure \ref{fig:pipeline}, we use Blender to render each source model and output a reference video. Then, we process all stimuli using the same views, generating images from these views using the rendering algorithm of the corresponding method and outputting it as a video. Specifically, we choose 3 view-to-object distances \(d_0=1.2m\), \(d_1=1.5m\), \(d_2=1.8m\) to render, and define the view-to-object distance \(d(\theta)\) as a function of the view rotation angle \(\theta \in [0^\circ, 1080^\circ]\):
\begin{equation}
d(\theta) \;=\; d_0 
+ (d_1 - d_0)\,\operatorname{tri}\!\Big(\tfrac{\theta}{360^\circ}\Big) 
+ (d_2 - d_1)\,\operatorname{tri}\!\Big(\tfrac{\theta-180^\circ}{720^\circ}\Big),
\end{equation}
where \(\operatorname{tri}(x) \;=\; 1 - \big|\,1 - 2\,(x - \lfloor x \rfloor)\,\big| \). Each video is 30 FPS and has 180 frames. In addition, each video has a uniform resolution of \(1080\times1080\). Note that since the input images used for reconstruction have no background, we manually add a gray background with RGB values of \((153, 153, 153)\) to each frame of the video.

\begin{figure}
    \centering
    \includegraphics[width=0.85\linewidth]{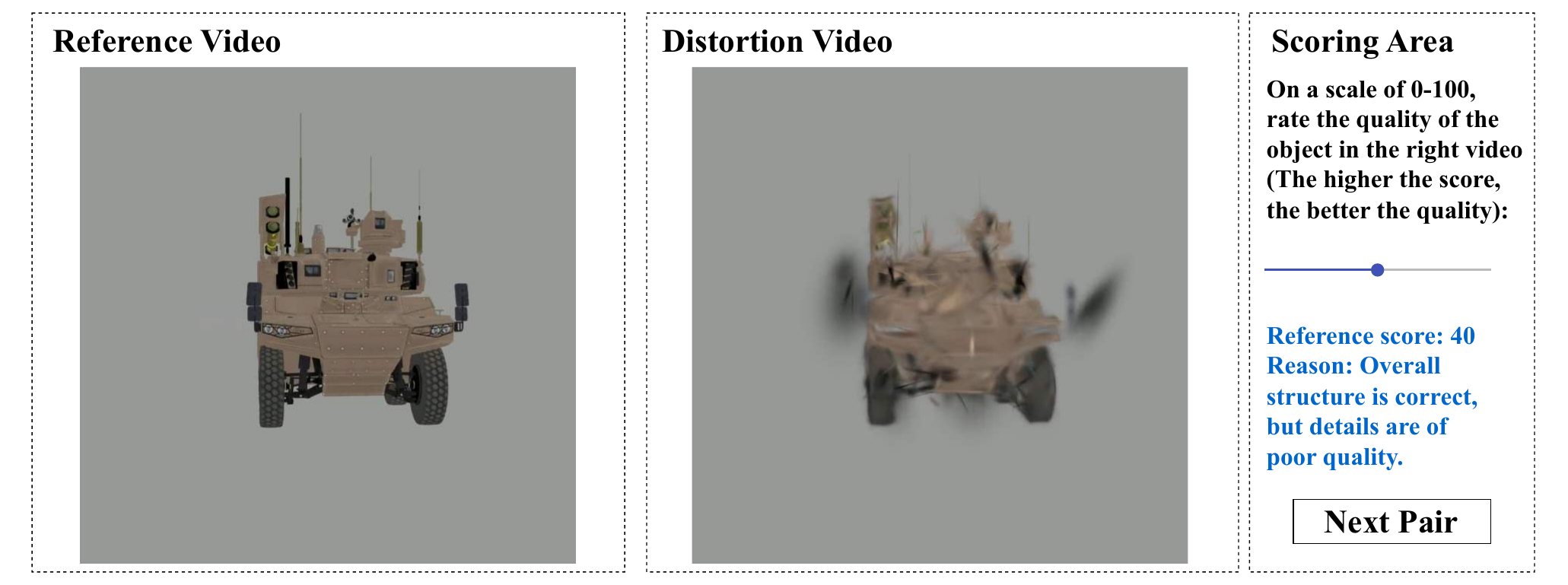}
    \caption{Scoring Interface.}
    \label{fig:scoring_interface}
\end{figure}

\textbf{Experiment.} To obtain reliable and controllable results \cite{jin2021just}, we start a crowdsourced project using MTurk \footnote{https://www.mturk.com} and create a scoring interface. As shown in Figure \ref{fig:scoring_interface}, the interface includes three modules: reference video, distortion video, and scoring area. After each pair of videos is played, workers are allowed to slide the scoring bar. In the training stage, a suggested score and a reason corresponding to the distortion will be displayed. After training, participants can enter the test stage of the experiment, during which the suggested scores will no longer be displayed, and the rest of the content remains the same as in the training stage. At the end of the experiment, the scoring results will be automatically uploaded, and after our review, the participants will be paid. Ultimately, 226,800 quality scores are collected and a total of 2,452 participants complete the experiment.

\section{Data Processing and Analysis}
\label{sec:analysis}

\subsection{Dataset Completion and Comparison}
To extract a sufficient and accurate set of valid scores, we adopt the following three-step filtering process. \textbf{(1) Filter by training stage scores:} If a participant's ranking of the scores of the three samples in the training stage does not match the order of the suggested scores, all scores of the current participant in this playlist will be filtered out. \textbf{(2) Filter by score distribution:} We refer to the ITU-R BT.500-13 screening procedure \cite{series2012methodology} to detect unreasonable score distributions. This procedure is the same as summarized in \cite{mantiuk2012comparison}. \textbf{(3) Filter by GUs:} Based on the Golden Units (GUs) in each playlist, we perform further score filtering. Unlike the approach in \cite{hossfeld2013best}, which filters after mapping to discrete values, we retain the original scores and filter according to the distribution of each score list.

\begin{table}[t]
    \centering
    \scriptsize
    \setlength{\tabcolsep}{2pt}
    \begin{tabular}{cccccccc}
    \toprule
    \textbf{Name} & 
    \textbf{Year} & 
    \textbf{Distortion Factor} &
    \textbf{SQA Views} &
    \textbf{\( N_{s} \)} & 
    \textbf{\( N_{o} \)} & 
    \textbf{\( N_{g} \)} & 
    \textbf{\( N_{m} \)} \\
    \midrule
    GSC-QA \cite{yang2024benchmark} & 2024 & Compression & $360^{\circ}$ & 9 & 6 & 120 & 1 \\
    NVS-QA \cite{zhang2025evaluating} & 2025 & / & $360^{\circ}$+Front & 13 & / & 65 & 3 \\
    GS-QA \cite{martin2025gs} & 2025 & / & $360^{\circ}$+Front & 8 & / & 64 & 7 \\
    3DGS-IEval-15K \cite{xing20253dgsieval15klargescaleimagequality} & 2025 & Compression & 20 Views & 10 & / & 760 & 6 \\
    MUGSQA (Ours) & 2025 & Input Settings & $1,080^{\circ}$ & / & 55 & 2,414 & 6 \\
    \bottomrule
    \end{tabular}
    \caption{Dataset Comparison. \( N_{s} \), \( N_{o} \), \( N_{g} \), \( N_{m} \) refer to the number of source scenes, source objects, labeled gaussians, and GS-based reconstruction methods, respectively.}
    \label{tab:dataset_comparison}
\end{table}

As a result, we retain 101,555 valid scores, ensuring that each sample in every playlist has at least 30 valid scores. Then we compute Mean Opinion Scores (MOS) by averaging the scores given by different participants on each stimulus. Similarly to \cite{yang2024benchmark}, we map the MOS to a continuous range of 0 to 5, where higher scores represent better quality. At this point, we have completed the dataset.

As shown in Table \ref{tab:dataset_comparison}, our dataset has several advantages over existing datasets. Firstly, MUGSQA compensates for the deficiencies in GT by using synthetic data. This not only includes the image data required for reconstruction, but also contains the 3D mesh models, providing more reliable comparisons and analyses. Secondly, MUGSQA addresses the shortcomings of existing datasets in single-object reconstruction. Most datasets only contain scenes, whereas our dataset encompasses 55 synthetic objects as source models. In fact, if a single object can be reconstructed, it will be more conducive to an in-depth analysis of the distortion characteristics and metric design. This need to assess the quality of a single Gaussian object is crucial in scenarios requiring a large number of high-quality synthetic objects \cite{ljungbergh2025r3d2realistic3dasset}. In terms of SQA methods, compared to other datasets that only render frames in a fixed scale, our dataset takes into account the quality differences generated by rendering at different scales, thereby achieving 180 rendered frames and covering as many as 3 cycles. In terms of data annotation, our subjective experiments are also more thorough, including 2,414 valid MOS.
\begin{table}[t]
    \centering
    \scriptsize
    \setlength{\tabcolsep}{2pt}
    \begin{tabular}{cccccccc}
    \toprule
    \textbf{Method} & 
    \textbf{\( R_{overall} \)} & 
    \textbf{\( R_{resolution} \)} & 
    \textbf{\( R_{quantity} \)} & 
    \textbf{\( R_{distance} \)} & 
    \textbf{\( R_{pc} \)} \\
    \midrule
    3DGS \cite{kerbl20233d} & 71.04 & \underline{73.04} & 69.32 & 66.17 & \underline{75.62} \\
    Mip-Splatting \cite{yu2024mip} & \textbf{73.06} & \textbf{73.77} & \textbf{71.70} & \textbf{68.35} & \textbf{78.42} \\
    LightGaussian \cite{fan2024lightgaussian} & \underline{71.42} & 72.95 & 70.83 & \underline{67.08} & 74.82 \\
    EAGLES \cite{girish2024eagles} & 70.41 & 70.70 & 70.88 & 64.87 & 75.20 \\
    Octree-GS \cite{ren2024octree} & 66.30 & 66.06 & 65.78 & 65.66 & 67.70 \\
    Scaffold-GS \cite{lu2024scaffold} & 63.41 & 60.56 & \underline{71.67} & 53.45 & 67.95 \\
    \bottomrule
    \end{tabular}
    \caption{Robustness Comparison on MUGSQA Dataset with Per-Column Best (Bold) and Second-best (Underlined) Values.}
    \label{tab:robustness_comparison}
\end{table}

\subsection{Robustness of GS-based Reconstruction Methods}
To evaluate the robustness of different GS-based reconstruction methods using the MUGSQA dataset, we define a robustness score \( R_{u} \in [0,100] \), which integrates three aspects: stability, consistency, and performance \cite{sun2025mfcqa}. Stability is derived from the coefficient of variation \( CV = \frac{\sigma}{\mu} \times 100\% \), consistency from the MOS range \( M = \max_i \{MOS_i\} - \min_i \{MOS_i\} \), and performance from the mean MOS \( \mu \). These are mapped to \([0,100]\) and combined as:
\begin{equation}
\begin{split}
& R_{u} =\ 0.4 \times \max(0,\ 100 - 2 \times CV)\ + \\
               & 0.3 \times \max(0,\ 100 - 20 \times M)\ + 
               0.3 \times \min(100,\ 10 \times \mu)
\end{split}.
\end{equation}
This score is computed for each \(u\) independently while keeping the others fixed, where \(u\) is the uncertainty settings introduced in Section \ref{sec:preparation}, so \( u \in \{resolution, quantity, distance, pc\} \). The final robustness \( R_{overall} \) is obtained by averaging \( R_{u} \) across different settings. As shown in Table \ref{tab:robustness_comparison}, Mip-Splatting achieves the highest \( R_{overall} \), while 3DGS, EAGLES and LightGaussian also show strong performance. However, Octree-GS and Scaffold-GS, designed for large-scene reconstruction, perform poorly in object reconstruction. We believe that optimizations in multi-scale rendering, as well as the coarse-to-fine training strategy, are key to improving the quality of Gaussian object reconstruction and the robustness of the algorithm. Conversely, some methods, such as Level-of-Detail (LOD), while having a more powerful upper limit in reconstruction range, will have their steps correspondingly affected when facing non-ideal input conditions, thereby leading to more severe distortion.

\subsection{Performance of Objective Quality Assessment Metrics}
Our dataset possesses rich 2D and 3D visual data, where quality can be assessed for different modalities. However, unlike data in point cloud or mesh formats \cite{yang2020predicting, liu2023point, nehme2023textured, cui2024sjtu}, quality assessment metrics specifically designed for the 3D modality of GS are still lacking. Therefore, we only use 2D metrics for benchmarking.

\begin{table}[htbp]
    \centering
    \scriptsize
    \setlength{\tabcolsep}{2pt}
    \begin{tabular}{lllcccccccc}
        \toprule
        & & \multicolumn{4}{c}{\textbf{Main Set}} & \multicolumn{4}{c}{\textbf{Additional Set}} \\
        \cmidrule(lr){3-6} \cmidrule(lr){7-10}
        & Metric & PLCC & SROCC & RMSE & KROCC & PLCC & SROCC & RMSE & KROCC \\
        \midrule
        & PSNR & 0.5848 & 0.5246 & 1.1026 & 0.3662 & 0.5146 & 0.4749 & 1.2873 & 0.3266 \\
        & PSNR-Y & 0.5865 & 0.5260 & 1.1009 & 0.3674 & 0.5264 & 0.4937 & 1.2765 & 0.3379 \\
        & SSIM & 0.3686 & 0.3695 & 1.2635 & 0.2496 & 0.3148 & 0.2724 & 1.4251 & 0.2023 \\
        & SSIM-C & 0.3660 & 0.3609 & 1.2650 & 0.2438 & 0.3161 & 0.2580 & 1.4244 & 0.1949 \\
        & FSIM & 0.6769 & 0.6776 & 1.0005 & 0.4787 & 0.5662 & 0.5774 & 1.2375 & 0.4030 \\
        & MS-SSIM & 0.6240 & 0.6354 & 1.0622 & 0.4448 & 0.5691 & 0.5769 & 1.2346 & 0.3995 \\
        & CW-SSIM & 0.7186 & \underline{0.7350} & 0.9453 & \underline{0.5304} & 0.7089 & 0.7235 & 1.0590 & 0.5306 \\
        & GMSD & 0.5176 & 0.5479 & 1.1630 & 0.3799 & 0.5955 & 0.5861 & 1.2062 & 0.3983 \\
        & NLPD & 0.5936 & 0.5947 & 1.0939 & 0.4137 & 0.5202 & 0.4990 & 1.2823 & 0.3454 \\
        & VSI & \underline{0.7209} & 0.7262 & \underline{0.9421} & 0.5252 & 0.6150 & 0.6248 & 1.1840 & 0.4371 \\
        & LPIPS-V & 0.4051 & 0.4090 & 1.2428 & 0.2769 & 0.5017 & 0.5081 & 1.2988 & 0.3690 \\
        & LPIPS-A & 0.4165 & 0.4428 & 1.2358 & 0.3009 & 0.5276 & 0.5455 & 1.2755 & 0.3893 \\
        & NIQE & 0.1656 & 0.0444 & 1.3405 & 0.0348 & 0.1777 & 0.1540 & 1.4775 & 0.1049 \\
        & PIQE & 0.0991 & 0.0126 & 1.3526 & 0.0088 & 0.2166 & 0.1838 & 1.4658 & 0.1261 \\
        & DBCNN & \textbf{0.8846} & \textbf{0.8800} & \textbf{0.6693} & \textbf{0.6927} & \textbf{0.9223} & \textbf{0.9075} & \textbf{0.6183} & \textbf{0.7301} \\
        & FID & 0.4782 & 0.5156 & 1.1938 & 0.3578 & \underline{0.7585} & \underline{0.7680} & \underline{0.9784} & \underline{0.5722} \\
        \bottomrule
    \end{tabular}
    \caption{Performance Comparison on MUGSQA Dataset with Per-Column Best (Bold) and Second-best (Underlined) Values. LPIPS-V refers to LPIPS (VGG), and LPIPS-A refers to LPIPS (AlexNet).}
    \label{tab:correlation}
\end{table}

\textbf{Metrics.} We select several representative Full-Reference (FR) and No-Reference (NR) Image Quality Assessment (IQA) metrics. Specifically, we select 12 FR metrics: PSNR, PSNR-Y, SSIM, SSIM-C, MS-SSIM, CW-SSIM, FSIM, GMSD, NLPD, VSI, LPIPS (VGG), LPIPS (AlexNet) \cite{zhang2018unreasonable}, and 4 NR metrics: NIQE, PIQE, DBCNN \cite{zhang2020blind}, FID. All these metrics are calculated using IQA-PyTorch\footnote{https://github.com/chaofengc/IQA-PyTorch}. It is worth noting that because some metrics are based on deep learning, their values are results computed after performing a five-fold cross-validation on the target dataset.

\textbf{Results and Evaluation.} For the results of each metric, if they are not within the specified MOS range, we use a four-parameter logistic regression to map them. Next, we calculate the correlation coefficients between each metric and MOS, including Pearson Linear Correlation Coefficient (PLCC), Spearman Rank-order Correlation Coefficient (SROCC), Root Mean Square Error (RMSE), and Kendall Rank Correlation Coefficient (KROCC). Table \ref{tab:correlation} shows the overall performance of each metric under the two subsets. Among the FR-IQA metrics, except for CW-SSIM and VSI, which perform relatively well, the rest of the metrics yield poor results, and even the LPIPS series, capable of extracting deep features, has difficulty distinguishing the quality of our dataset samples. There are many influencing factors, such as the presence of pure color or empty backgrounds affecting the calculation results of some metrics, the difficulty in distinguishing quality differences after sample filtering, and the features extracted by some metrics from pre-trained DNNs not aligning with the characteristics of GS distortion. These factors collectively lead to the deterioration of the correlation coefficient results of these IQA metrics. For NR-IQA metrics, traditional NIQE and PIQE metrics perform very poorly, clearly indicating that their calculation methods are not suitable for assessing the quality of Gaussian objects. For the more advanced metric, DBCNN, it is able to achieve good results after fine-tuning. This demonstrates the importance of deep learning in modern quality assessment and the powerful ability of these architectures for fine-grained distinction. Based on these results, we find that the IQA metrics that only use 2D rendering results are not sufficient to evaluate the quality of Gaussian objects. Therefore, we call for the design of new metrics specifically for the GS modality, and we believe that if novel metrics can be further optimized to handle the Gaussian attribute designs of different methods, the effectiveness and speed of GSQA will reach even higher levels.

\section{Conclusion}
\label{sec:conclusion}
In this paper, we propose a unified multi-distance SQA method. Based on this, we construct a large-scale Gaussian object reconstruction dataset, MUGSQA, and establish two brand new benchmarks through SQA experiment, post-analysis, and filtering. By evaluating the performance of various GS-based reconstruction methods and various existng metrics on this benchmark, we believe that designing new GSQA metrics and conducting a deeper distortion analysis from a multi-modal perspective is an urgent need.

\section{Acknowledgement}
\label{sec:acknowledgement}
This research is partially supported by the Ministry of Education, Singapore, under the funding of MOE-T2EP20123-0006.

\vfill\pagebreak

\bibliographystyle{IEEEbib}
\bibliography{main}

\end{document}